\documentclass{article}

\usepackage{amsmath,amssymb}
\usepackage{bm}

\usepackage[preprint]{neurips_2026}

\usepackage{wrapfig}
\usepackage[utf8]{inputenc} % allow utf-8 input
\usepackage[T1]{fontenc}    % use 8-bit T1 fonts
\usepackage[breaklinks=true]{hyperref}       % hyperlinks
\usepackage{url}            % simple URL typesetting
\usepackage{booktabs}       % professional-quality tables
\usepackage{amsfonts}       % blackboard math symbols
\usepackage{nicefrac}       % compact symbols for 1/2, etc.
\usepackage{microtype}      % microtypography
\usepackage{xcolor}         % colors
\usepackage{multirow}
\usepackage{algorithm}
\usepackage{algpseudocode}
\usepackage{graphicx}
\usepackage{subcaption}

\definecolor{darkgreen}{RGB}{0,90,60}

\title{Plan First, Diffuse Later: Extrinsic Graph Guidance for Long-Horizon Diffusion Planning
}

\author{%
  Yaniv Hassidof\thanks{Corresponding author. Email: yaniv\_hass@campus.technion.ac.il} \\
  Technion \\
  \And
  Adir Morgan \\
  Technion \\
  \And
  Yilun Du \\
  Harvard \\
  \And
  Kiril Solovey \\
  Technion \\
}

\begin{document}

\maketitle
% \begin{abstract}
\begin{abstract}
Compositional diffusion models offer a promising route to long-horizon planning by denoising multiple overlapping sub-trajectories while ensuring that together they constitute a global solution.
However, enforcing local behavior over long chains is often insufficient for a coherent global structure to emerge.
Recent works tackle this limitation through \emph{intrinsic} search, which explores multiple paths \emph{during} the denoising process.
While intrinsic search improves global 
coherence, it comes at the cost of repeated evaluations of an already compute-heavy model. In this work, we argue that \emph{extrinsic} search, performed \emph{outside} the denoising process, offers a more effective mode of exploration for long-horizon planning while naturally enabling the use of classical algorithms to solve unseen combinatorial tasks at test time. 
Our eXtrinsic search-guided Diffuser (XDiffuser)
first computes a plan over a state-space graph---serving as a lightweight local connectivity oracle for the diffusion model. The plan is then used to guide denoising for a \emph{single} trajectory, effectively offloading the burden of exploration. 
XDiffuser outperforms diffusion-based baselines on long-horizon tasks, with particularly large gains in the low-quality data regime and on unseen tasks beyond goal-reaching, including multi-agent coordination and TSP-style reasoning. Website: 
\url{https://yanivhass.github.io/XDiffuser-site/}
\end{abstract}

\begin{figure}[ht]
    \centering
    \vspace{-1.0em}
    \begin{subfigure}[t]{0.31\textwidth}
        \centering
        \includegraphics[width=\linewidth]{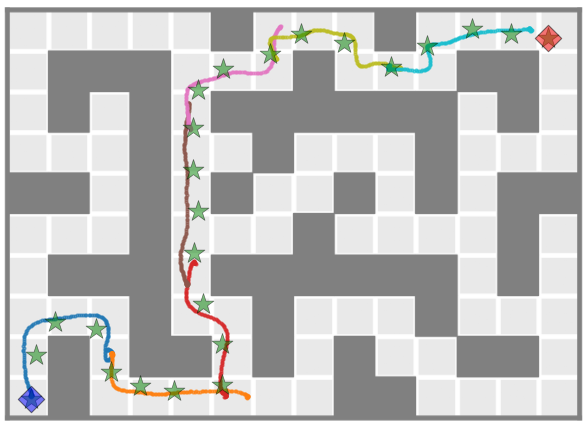}
        % \caption{Goal reaching.}
        \label{fig:maze_trajectory}
    \end{subfigure}
    \hfill
    \begin{subfigure}[t]{0.23\textwidth}
        \centering
        \includegraphics[width=\linewidth]{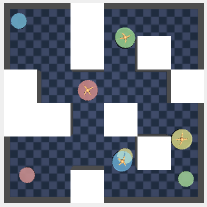}
        % \caption{Multi-agent planning.}
        \label{fig:ant_mapf}
    \end{subfigure}
    \hfill
    \begin{subfigure}[t]{0.30\textwidth}
        \centering
        \includegraphics[width=\linewidth]{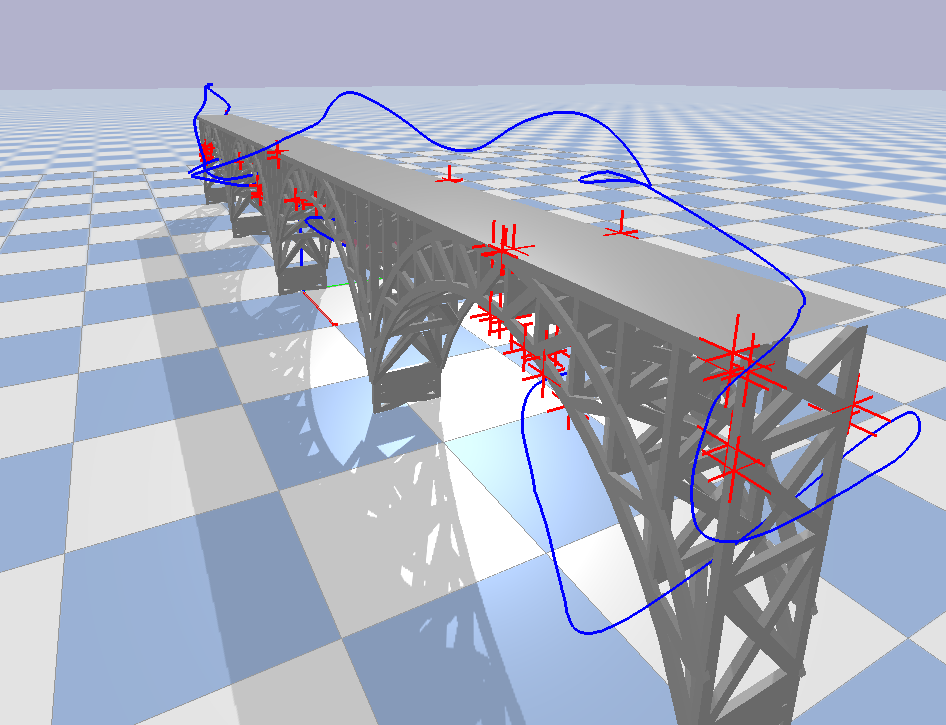}
        % \caption{Inspection planning.}
        \label{fig:bridge_inspection}
    \end{subfigure}
    
    \caption{By leveraging task-specific graph-search mechanisms, XDiffuser enables a pretrained goal-reaching compositional diffuser to solve complex unseen tasks. \textbf{Left:} Waypoints endow local segments with a coherent global structure, strengthening long-horizon goal reaching. The agent is marked by {\color{blue}$\blacklozenge$}, waypoints by {\color{darkgreen}$\bigstar$}, and the goal by {\color{red}$\blacklozenge$}. \textbf{Middle:} Multi-agent goal-reaching AntMaze. \textbf{Right:} XDiffuser performs an inspection planning task when paired with an existing graph-based solver. Points of interest for inspection are marked in red, XDiffuser's trajectory shown in blue.}
    % \kiril{If space becomes an issue, embed the subcaption in a white text box within each figure and remove the subcaption}
    % Planning tasks evaluated in our experiments. Left: The agent represented by the orange ball plans a trajectory to the goal marked with a red star, using diffusion guided by waypoins marked as green stars. Middle: a multi agent planning scenario, where agents must reach their goal while avoiding collision with the others. Right: trajectory generated by XDiffuser for a bridge inspection  task featured in experiment~\ref{ip_experiment}. 
    % }
    \label{fig:experimental_settings}
    \vspace{-0.5em}
\end{figure}

\section{Introduction}

Learning-based planning is pushing the frontiers of robotic decision-making, extracting effective behavior from static datasets collected by short-horizon, often suboptimal policies. However, scaling these methods to long-horizon tasks such as robotic assembly~\citep{JIANG2022102366}, navigation~\citep{nahavandi2025comprehensive} and infrastructure inspection~\citep{inspection} remains a central challenge~\citep{park2025horizon}, as both learned dynamics models~\citep{moerland2023model} and regressed value-based policies~\citep{model_free} suffer from compounding approximation errors.

% trajectory generators, such as diffusion planners~\citep{janner2022diffuser}, that excel at the segment level frequently fail to produce globally consistent plans. 

Diffusion-based planners~\citep{janner2022diffuser} offer an appealing alternative by learning to directly sample a coherent trajectory from the data distribution, thereby casting planning as inference~\citep{botvinick2012planning}. 
% This perspective has led to strong short-horizon performance, especially in settings where planning must proceed based on static demonstrations. \kiril{can you give examples of problem instances that are now considered "solved" using those methods? Otherwise, can you be more specific about what is meant by "impressive"?}
Yet, planning-as-inference faces its own limitations in the long-horizon regime. Real-world demonstrations required for training are often short, local, and incomplete due to the monetary, time, and safety constraints of data collection, thus placing long-horizon problems inherently outside the data distribution. 
%while successful long-horizon behavior requires composing many such segments into a globally coherent solution.
To address this limitation, recent approaches leverage \emph{compositionality}~\citep{mishra2023generative_corll,luogenerative} to assemble short-horizon trajectory segments that are mutually compatible into a coherent global long-horizon solution. 
%at test time, it must reason over trajectories that exceed the training horizon, placing it inherently outside the data distribution. T Recent works take an important step toward compositionality by learning to generate short-horizon trajectory fragments that are mutually consistent and compatible. 
% However, as the horizon grows, local segments must balance local quality against compatibility with an increatheir neighbors, inducing competing objectives across the plan
However, as the horizon grows, compatibility must be maintained over an increasingly long chain of segments, each attempting to maximize its own local likelihood, making the inference process increasingly brittle: the model may generate locally plausible segments that are mutually incompatible, leading to failures in global consistency, task completion, or constraint satisfaction.
% ---for example, a locally high-quality segment may be difficult to connect to adjacent fragments. Consequently, 

\begin{figure}[h]
    \centering
    \includegraphics[width=0.75\linewidth]{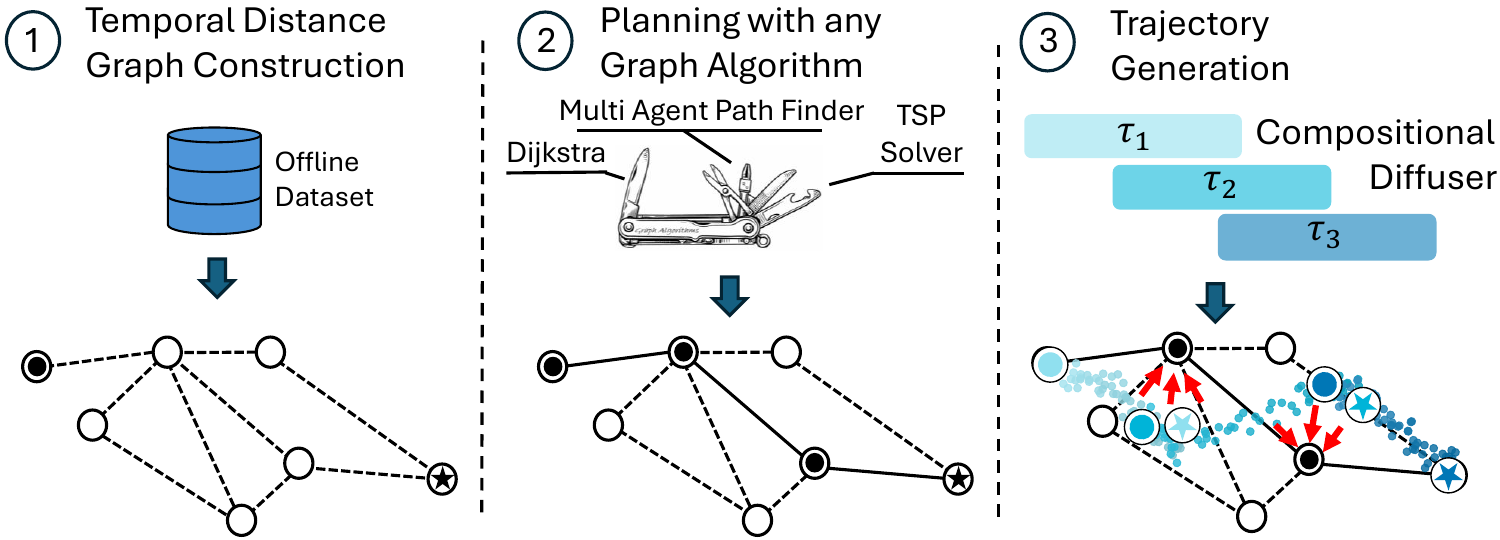}
    \caption{XDiffuser decomposes planning into \textit{extrinsic} search followed by guided \textit{intrinsic} generation. \textbf{(1)} At training time, a temporal distance representation is used to construct a connectivity graph over sampled dataset states. \textbf{(2)} A task-appropriate graph search is executed, producing a sequence of waypoints representing the graph solution. \textbf{(3)} A pretrained CompDiffuser denoises a smooth trajectory, guided by the waypoints set. }
    \label{fig:method}
    \vspace{-1.0em}

\end{figure}

We argue that this limitation is not merely a consequence of approximation error during inference, but a fundamental structural limitation of pure planning-as-inference approaches: when the target long-horizon distribution is never observed, local consistency alone is insufficient to guarantee globally-coherent behavior. Our key insight is that these requirements are naturally handled by \emph{extrinsic} search (i.e., \emph{outside} of denoising), while diffusion is best suited to synthesizing smooth, dynamically plausible trajectories between compatible states. Based on this view, we introduce %propose to separate global coordination from local generation Our 
the eXtrinsic search-guided Diffuser (XDiffuser), which is visualized in Figure \ref{fig:method}. Concretely, at training time, XDiffuser builds a graph over real trajectory data to produce a sparse, globally consistent scaffold. At test time, XDiffuser \emph{first} searches this graph scaffold to produce a high-level plan that is \emph{then} used for guiding 
a pretrained compositional diffusion planner %is then guided by this scaffold via graph search 
to generate continuous trajectories that are smooth, likely under the data, and executable. In contrast to prior approaches that perform search \emph{inside} the denoising process~\citep{zhang2025inference,mishra2026cdgs,yoon2025cmctd}, %, our method uses search outside the generative loop. 
XDiffuser avoids repeatedly querying the model during iterative denoising, making planning substantially more efficient; we discuss this distinction in more detail in the related work. %\kiril{this paragraph requires TLC and making some aspects more precise. the algorithm name needs to be introduced here as well.}

This decomposition between search and denoising yields a simple and modular framework for long-horizon planning. As the search layer operates over a structured graph it naturally supports adaptive horizon selection: the diffusion horizon is inferred from the temporal duration of the selected route, rather than fixed in advance by a heuristic rollout length as in prior methods. 
Furthermore, since the high-level objective lives in the search procedure,
% rather than in retraining the generative model, new costs, constraints, and even 
different planning algorithms can be incorporated at test time, enabling extensions beyond standard goal-reaching to settings such as multi-agent coordination or inspection planning (Figure \ref{fig:experimental_settings}).

% \kiril{this is essentially a repetition of the previous paragraph. Instead, we need to discuss the experiemntal results.} Our contribution is threefold. First, we introduce a structured search procedure over the training data that provides a globally consistent scaffold and guidance for compositional diffusion planning. Second, we show that this design enables adaptive-horizon planning, allowing the system to solve tasks whose effective horizon exceeds what was seen during training. Third, we demonstrate controllable generalization through graph construction and edge constraints, making explicit the tradeoff between staying close to the dataset and exploring novel long-range compositions. Finally, we show that the framework is readily extendable to new test-time objectives and constraints, yielding a versatile recipe for long-horizon robotic planning.
% \YH{new closing paragraph, replaced the old one}
Our experimental results (Section~\ref{experiments}) highlight XDiffuser’s test-time scalability. On the OGBench suite~\citep{ogbench_park2025}, XDiffuser exhibits strong long-horizon planning, especially after training on low-quality data, by achieving 98.5\% success rate on AntMaze Large Explore---over 70\% higher than its base diffusion planner. 
We then demonstrate strong generalization across task structure while reusing the same pretrained diffuser in two tasks:  (i) a multi-agent setting, where the single-agent diffuser is paired with priority-based graph planner to enable coordinated behavior, and (ii) inspection planning, leveraging a TSP-style graph planner.

\section{Related Work}
\label{related_work}

\textbf{Planning with Graphs.}
Planning over explicit graphs is one of the oldest, most widespread paradigms for long-horizon decision making. In robot motion planning, search is often carried out on a graph~\citep{cohen2011planning,kavraki1996probabilistic,panasoff2025effective}, capturing the problem's state-space connectivity, to generate efficient collision-free paths. Moreover, state-space graphs are leveraged in obtaining complex behaviors beyond standard goal reaching, such as multi-robot coordination~\citep{Wang2025WherePC} or inspection planning~\citep{morgan2026scalable}. 

Graph search has been widely combined with reinforcement learning (RL) for long-horizon planning. Prior work plans over graphs of observations or learned landmarks, using image-space graphs~\citep{savinov2018semi, liu2020hallucinative}, value-guided subgoal search~\citep{eysenbach2019search}, or latent landmarks with multi-step edges~\citep{zhang2021world}. More recently, Graph-Assisted Stitching (GAS)~\citep{baek2025graph} performs search in a latent temporal-distance space. These approaches decompose long-horizon problems into locally feasible steps, but rely on strong value-based policies to track graph waypoints.
However, this abstraction is inherently limited. Graph edges provide only local, approximate feasibility—indicating that transitions are possible or observed—without guaranteeing that sequences of edges form likely trajectories under the data distribution. As a result, individually valid transitions may compose into globally unnatural behaviors~\citep{zhang2021world}, and shortest-path objectives can favor brittle plans that cut corners or traverse rarely co-occurring states.

To address these shortcomings, our graph-guided diffusion approach takes inspiration from robot autonomy pipelines and hierarchical motion planning approaches wherein a high-level graph plan is used to invoke a mid-level trajectory optimization approach to produce long-horizon dynamically feasible trajectories~\citep{betz2023tum,wahba2024kinodynamic,huang2025safe}. 
% classical motion planning, where a popular method \YH{(I need examples here)} to overcome this limitation is by combining high-level graph search with mid-level trajectory optimization (e.g., TrajOpt~\citep{schulman2014motion}).
% This combination is the inspiration for our Graph-guided diffusion approach. 
However, in our setting, rather than refining trajectories only via gradients of hand-designed costs and assumed-known obstacles, the diffusion model leverages a learned score function to synthesize trajectories that are smooth, dynamically feasible, and consistent with the data distribution.

% Our approach addresses this limitation \kiril{which one?} by decoupling high-level planning from trajectory generation. We use graph search to produce a coarse sequence of waypoints, but defer the construction of the final trajectory to a diffusion-based planner conditioned on this guidance. 
% This way, the graph provides a backbone for efficient search and global structure, while the diffusion model ensures coherence and robustness, without inheriting the brittleness of purely discrete plans.
% This structure is reminiscent of classical motion planning pipelines that combine global search with local trajectory optimization (e.g., TrajOpt~\citep{schulman2014motion}). However, rather than refining paths via the gradient of hand-designed costs and constraints, the diffusion model leverages the learned score function to synthesize trajectories that are smooth, dynamically feasible, and consistent with the data distribution.

% This combination retains the efficient exploration and compositionality of graph search while avoiding the brittleness of purely discrete plans.

\textbf{Planning with Diffusion.}
Diffusion models~\citep{ho2020denoising} have emerged as expressive trajectory priors and planners. 
% \YH{mention guided diffusion and how the both translate nicely to trajectory generation as shown in,,,} 
Diffuser~\citep{janner2022diffuser} formulates planning as iterative denoising over entire trajectories, 
% while demonstrating their flexibility through various guidance signals, both 
achieving strong performance while enabling controllability through guidance~\citep{dhariwal2021diffusion}, yet remaining limited to the horizon lengths observed during training.
% , and Diffusion Planner~\citep{zheng2025diffusion} demonstrates the strength of this view in autonomous driving. 
% Subsequent work has tried to overcome the fixed-horizon limitations of trajectory diffusion by generating a variable count of overlapping subsequent segments, forming together a longer horizon trajectory. 
% Diffusion Forcing~\citep{chen2024diffusion} combines next-token prediction with full-sequence diffusion and supports variable-horizon rollouts beyond the training horizon. 
Generative Skill Chaining (GSC)~\citep{mishra2023generative_corll} and CompDiffuser~\citep{luogenerative} aim to synthesize long-horizon behavior from short demonstrations by stitching overlapping denoised segments. %This line of work paves a path toward planning at horizons not seen during training. 
In practice, however, maintaining global structure over long horizons remains difficult as denoising of each segment only follows a local consistency objective.  
% : 
% denoised segments face competing objectives---local coherence vs. strong overlap with potentially far away segments \kiril{the last part is unclear}. 
A common failure case of such methods is sampling of neighboring segments from incompatible modes of the trajectory distribution, leading to invalid states at segment overlaps due to mode averaging~\citep{mishra2026cdgs}. 
% neighboring segments can induce competing objectives, and small local ambiguities can accumulate into globally incoherent trajectories.

% A recent line of work attempts to guide long-horizon stitching by injecting search directly into diffusion inference. 
% Compositional Monte Carlo Tree Diffusion (C-MCTD)~\citep{yoon2025cmctd} performs tree search by treating partial plans as nodes, and possible extensions (via stitching of an additional overlapping segment) as edges.   
% Monte Carlo Tree Diffusion (MCTD)~\citep{yoon2025mctd} performs tree search over partially denoised trajectory segments, and C-MCTD~\citep{yoon2025cmctd} extends this idea to compose longer plans from shorter ones. 
% Compositional Monte Carlo Tree Diffusion (C-MCTD)~\citep{yoon2025cmctd} performs tree search using partially denoised trajectory segments as edges, connected by a shared overlapping state. 
% DiTree \cite{hassidof2025ditree} similarly combines diffusion priors with state-space tree expansion for kinodynamic motion planning. 
% More generally, 
% Inference-Time Scaling of Diffusion Models through Classical Search~\citep{zhang2025inference} and 

A recent line of work attempts to strengthen compositionality by injecting search directly into diffusion inference to guide long-horizon stitching~\citep{zhang2025inference, yoon2025cmctd,mishra2026cdgs}, iteratively calling the diffusion denoiser to search over possible denoising steps. 
These methods show that additional test-time compute can substantially improve diffusion planning, but they also reveal a key bottleneck: when search is placed \emph{inside} denoising, the computational cost scales with both search complexity and diffusion depth. In the worst case, a search procedure with branching factor $b$, depth $H$, and $K$ denoising steps per proposal requires
% $
% O\!\left(K \sum_{d=0}^{H-1} b^d\right)=O(K b^H)
% $
$
O(K b^H)
$
denoiser calls~\citep{russell2016artificial}.
Compositional Diffusion with Guided Search (CDGS)~\citep{mishra2026cdgs} reduces this cost by maintaining a fixed-sized population of candidate plans that is iteratively resampled and pruned, but by doing so, it sacrifices explicit backtrack-capable search structure.

% Compositional Monte-Carlo Tree Diffusion (C-MCTD)~\citep{yoon2025cmctd}, performs tree search via an \textit{online composer}, in which nodes represent partial trajectories and edges correspond to possible plan extensions. 
% C-MCTD attempts to circumvent the test-time denoising burden by proposing an additional \textit{preplan composer}---a graph built at training time where edges between dataset states are generated via calls to the \textit{online composer}. 
Compositional Monte-Carlo Tree Diffusion (C-MCTD)~\citep{yoon2025cmctd} performs tree search via an \textit{online composer}, where nodes represent partial trajectories and edges correspond to candidate plan extensions.
To reduce the burden of test-time denoising, C-MCTD introduces a \textit{preplan composer}: a graph constructed during training, in which edges between dataset states are generated by querying the online composer.
At inference, start and goal are connected to this graph, and shortest-path search returns a sequence of edges that are stitched into the final trajectory.
However, constructing this graph requires running the full search procedure between many state pairs, leading to a worst-case cost of $O(V^2K b^H)$ denoiser calls for $V$ vertices. %As mentioned by the authors of C-MCT thus difficult to scale, which the authors mention 
% This renders the preplanner inapplicable for some of their tested environments. 
Moreover, edges only approximately match endpoint positions (within $\epsilon$) and ignore other state variables (e.g., velocity, orientation), which can introduce inconsistent transitions. %\YH{C-MCTD is the method most related to ours, but we might be dedicating \textit{too much} text for it, what do you think?}. \kiril{I'm fine with the way it is}
% immediate execution of the retreived trajectory  ignoring discontinuities and infeasible transitions \kiril{elaborate}. 
In contrast, XDiffuser generates a single coherent and dynamically feasible trajectory via waypoint-guided diffusion. This distinction enables us to perform large-scale search, as we show in multi-agent path finding and inspection planning tasks, while successfully handling complex dynamics. %\kiril{it's unclear whether this sentence contrasts with all the previous papers in this section, or just C-MCTD. If it's the latter, then it's inaccurate}
% \YH{Furthermore, scaling such a graph incurs high computational cost, leading to relatively sparse graphs. The authors mention this renders the graph ineffective for high dimensional tasks such as manipulation. (which hopefully we will be able to show)}

In different settings, several recent works introduce a high-level discrete scaffold precisely because pure trajectory denoising struggles with long-horizon discrete decision making: DiTree~\citep{hassidof2025ditree} combines a diffusion policy with state-space tree expansion for kinodynamic motion planning. DGD~\citep{liang2025dgd} uses discrete multi-agent paths to guide continuous diffusion in multi-robot planning, DiMSam~\citep{fang2024dimsam} uses task-and-motion planning to compose learned diffusion samplers, and Hybrid Diffusion~\citep{hoeg2025hybrid} jointly models symbolic and continuous plans. In contrast to those works, our XDiffuser performs search outside of denoising %Our perspective is complementary: rather than spending search budget inside the denoising process, we use graph search 
using a lightweight and general-purpose planning scaffold. %, and does not require the short-horizon diffusion model by itself to discover a globally successful long-horizon trajectory. 

\section{Problem Formulation}
\label{prob_formulatiom}
We consider the problem of offline long-horizon motion planning for a robot operating in a continuous state space. We are given a fixed dataset $\mathcal{D}$ of previously collected trajectories, where no further interaction with the environment is allowed prior to deployment. Each trajectory is a sequence $\tau = (s_1, a_1, \dots, s_T, a_T)$, where $s_t \in \mathcal{S}$ denotes the robot’s state (e.g., position, velocity, orientation) and $a_t \in \mathcal{A}$ denotes the control input. In practice, our method requires only access to states, and actions are used to train an inverse dynamics model or derived by a controller, in accordance with existing methods~\citep{ajay2023is,luogenerative}. Importantly, the dataset consists of short-horizon behaviors collected under one or more behavior policies and does not necessarily contain trajectories that solve the long-horizon tasks of interest. Accordingly, models are trained on local windows extracted from $\mathcal{D}$, each of length at most $H_{\mathrm{train}}$, and thus only capture short-term, local behavior rather than complete task solutions.

At test time, the robot is given an initial state $s_{\mathrm{start}}$ and a task specification, such as a goal state, cost function, or additional constraints. The objective is to generate a feasible trajectory $(s_1, \dots, s_{H_{\mathrm{test}}})$ that satisfies the task, where typically $H_{\mathrm{test}} \gg H_{\mathrm{train}}$. %This trajectory is then executed by a downstream controller (e.g., a tracking controller or inverse dynamics model) that converts the planned state sequence into control actions. \kiril{haven't we already discussed this above?}
The central challenge is therefore to compose locally valid behaviors into a globally consistent, long-horizon plan.
% despite only observing short segments during training.

% Our approach addresses this setting by combining two components: (i) a diffusion-based planner trained on short windows from $\mathcal{D}$ to model locally likely transitions, and (ii) a graph constructed over states in $\mathcal{D}$ to provide global structure at test time. \adir{Maybe worth switching to "graph constructor" - graph is the product of a graph constructor algorithm we propose, not directly our component}
% Performing a search over this graph yields a coarse sequence of waypoints grounded in real data, which guides a diffusion process that generates smooth, dynamically consistent trajectories between them. 

\section{Method}
\label{method}
We describe our XDiffuser approach 
%We propose \emph{graph-guided compositional diffusion} 
for long-horizon planning from offline data. The key idea is to separate \emph{global coordination} from \emph{local trajectory generation}. Rather than relying solely on overlap consistency to propagate information across a long chain of locally generated segments, XDiffuser leverages a connectivity graph to first search for a coarse, globally consistent sequence of temporal waypoints. XDiffuser then uses this waypoint scaffold as a soft energy term defined over the full trajectory to guide compositional diffusion toward a coherent global solution. 
Overall, each segment is generated under three complementary constraints: local feasibility from the pretrained diffusion prior, consistency with adjacent segments through learned overlap coupling, and global coordination through the waypoint scaffold. We now detail each component of our method.

% This preserves the local coupling structure of compositional diffusion while encouraging the full trajectory to follow a coherent global route. 
% As a result, each segment is generated to be simultaneously consistent with its neighbors and with the high-level plan, reducing the effective long-range dependency of the sampling problem.

\subsection{Connectivity Graph Construction}
\label{sec:graph_construction}

We now describe our construction of an undirected graph 
\(\mathcal G = (\mathcal V, \mathcal E)\) 
over the offline dataset to capture coarse, task-agnostic connectivity between states, as illustrated in Algo~\ref{algo1}.
We first uniformly sample \(N\) states from the dataset $\mathcal{D}$ to form the vertex set
$\mathcal V = \{v_1, \dots, v_N\} \subset \mathcal S.$
To determine connectivity, we rely on a learned temporal-distance representation (TDR) 
\(f_\psi : \mathcal S \to \mathbb R^d\) as formulated by~\citep{park2024foundation,baek2025graph}. 
In this representation, Euclidean distance reflects temporal proximity between states.
Importantly, TDR is used only to define distances and neighborhood structure, whereas the graph vertices lie in the original state space. 
We define the edge cost as
$c(v_i, v_j) = \|f_\psi(v_i) - f_\psi(v_j)\|_2.$
Each vertex \(v_i\) is connected to its \(k\) nearest neighbors under this cost, 
subject to a connectivity threshold 
% \(\tau\):
% $(v_i, v_j) \in \mathcal E 
% \quad \text{if} \quad 
% v_j \in \mathrm{kNN}(v_i) \;\; \text{and} \;\; 
$\|f_\psi(v_i) - f_\psi(v_j)\|_2 \le \alpha$, which we adopt from~\citep{baek2025graph}. 
% which we set as the target temporal distance $H_{TD}$ described by~\citep{baek2025graph}.
% To ensure global connectivity, we retain only the largest connected component of the resulting graph, 
%discarding isolated vertices. \kiril{isn't this problematic? who said that a given query must be solved using the largest connected component?}
At test time, task-dependent states (e.g., start, goal) are added to the graph using the $\mathrm{kNN}$ procedure.

\begin{wrapfigure}{r}{0.48\textwidth}
\vspace{-2.0em}
\begin{minipage}{0.48\textwidth}
\begin{algorithm}[H]
\caption{Connectivity Graph Construction}
\begin{algorithmic}[1]
\State Sample \(N\) states \(\mathcal V \subset \mathcal D\)
\State Compute TDR  \(z_i = f_\psi(v_i), \forall v_i\in \mathcal{V}\) 
\State \(\mathcal E \leftarrow \emptyset\)
% \Statex
\State \textbf{for} each \(v_i \in \mathcal V\) \textbf{do}
\State \quad \(\mathcal N_i \leftarrow \mathrm{k}\)NN of \(z_i\)
\State \quad \textbf{for} each \(z_j \in \mathcal N_i\) \textbf{do}
\State \quad\quad \textbf{if} \(\|z_i - z_j\|_2 \le \alpha\) insert \((v_i,v_j)\) to \(\mathcal E\)
% \State Keep largest connected component of \((\mathcal V,\mathcal E)\)
\end{algorithmic}
\label{algo1}
\end{algorithm}
\end{minipage}
\vspace{-1em}
\end{wrapfigure}

% This construction is intentionally simple and avoids additional clustering or structure. 
The graph $\mathcal{G}$ encodes only coarse reachability, i.e., which states can plausibly connect, while deferring the 
generation of smooth and dynamically consistent trajectories to the downstream diffusion model. 
As a result, the graph is scalable, reusable across tasks, and independent of the final planning objective. 
While more sophisticated graph construction methods could be incorporated~\citep{zhang2021world,baek2025graph}, 
we find this minimal design sufficient for guiding a long-horizon diffusion planner.

\vspace{-0.5em}
\subsection{Planning via Off-the-Shelf Graph Algorithms}
\label{sec:graph_planning}
Next, the graph \(\mathcal G\) is used  to produce waypoints for downstream diffusion guidance. 
At test time, we are given a planning problem specified by an initial state \(s_{\mathrm{start}}\) and a task objective, such as a goal state or constraint. We invoke a graph search algorithm to find a plan that satisfies those constraints. % We will now detail \kiril{seems like some text was accidentally deleted here} begin by finding its graph solution. 
For example, in a goal-reaching task, a shortest path $\tau_{\mathcal G}:=(v_0 = s_{\mathrm{start}}, v_1, \dots, v_R = s_{\mathrm{goal}})$ over \(\mathcal G\) can be obtained, as illustrated in Fig~\ref{fig:graph_path}.
Since graph edge costs approximate temporal distance, the path induces a nominal timing along the route through the cumulative costs
$t_r = \sum_{i=0}^{r-1} c(v_i,v_{i+1})$.
However, a dense sequence of waypoints is often too restrictive for downstream diffusion of a smooth trajectory. Thus, we downsample this path $\tau_{\mathcal G}$ at a fixed temporal interval \(\Delta t\) to obtain the final sequence of temporal waypoints
$\mathcal W = \{(w_m,\hat t_m)\}_{m=0}^{M}$,
where \(w_m\) is the selected graph state and \(\hat t_m\) is its nominal time along the route. The interval \(\Delta t\) determines the density of the waypoint scaffold: smaller values produce tighter control over the diffusion process, at the cost of possibly over-constraining the diffusion model to produce infeasible trajectories. 
% \YH{An important aspect is graph paths favor optimality over dynamics feasibility, i.e. the nominal timing often represents infeasible timing. To enhance feasibility, we dilate our nominal timing such that the number of generates chunks roughly matches the number chosen in~\citep{luogenerative}. }
An important observation is that graph shortest paths prioritize global optimality at the cost of possibly inducing dynamically infeasible nominal timing. 
% Consequently, the nominal timing associated with these paths often underestimates the true traversal time. 
To better align the plan with downstream diffusion, we temporally dilate our waypoints by a constant factor so that the number of generated chunks roughly matches the number of chunks reported by~\citet{luogenerative}.
% while larger values provide a coarser route description.

% \adir{I think we can organize it so this clarification is not needed.}
% Importantly, shortest-path search is only one instantiation of this interface. Our method only requires the graph layer to return an ordered sequence of vertices and extract their respective nominal timing. Any graph algorithm with this output format can be used instead, including prioritized multi-agent path finding, coverage planning, or traveling-salesman-style routing. The diffusion model is kept fixed; only the high-level graph objective changes. \adir{examples are slightly repeated.}

\subsection{Waypoint-Guided Compositional Diffusion}
\label{sec:waypoint_diffusion}

We now explain how the graph solution, captured by the temporal waypoints $\mathcal W$, is used as a coarse scaffold for long-horizon trajectory generation. 
Let \(\tau = (s_1,\dots,s_H)\) denote the full trajectory to be computed using the diffusion model. 
Following~\citet{luogenerative}, \(\tau\) can be decomposed  into \(K\) segments 
\(\tau_1,\dots,\tau_K\) with $\mathcal{O}$ overlapping states between every subsequent pair, represented by the approximated distribution
\[
p_\theta(\tau \mid q_s, q_g) \propto
p_\theta(\tau_1 \mid q_s, \tau_2)\,
p_\theta(\tau_K \mid \tau_{K-1}, q_g)\,
\prod_{k=2}^{K-1}
p_\theta(\tau_k \mid \tau_{k-1}, \tau_{k+1}),
\]
where every factor $p_\theta$ is generated by the same short-horizon diffusion model with weights $\theta$, enabling long-horizon generation by enforcing local consistencies while neglecting non-local dependencies~\citep{yedidia2005constructing}. 
% However, for long horizons, global consistency must propagate through a chain of overlapping local constraints, which can become brittle. 

To inject the missing global structure, we leverage the waypoint sequence $\mathcal W$ via gradient-based diffusion guidance~\citep{carvalho2023motion, song2023lossguided}. We begin by initializing the noisy trajectory prior to denoising by treating the nominal time of the final waypoint $w_M$ as the expected goal-reaching time.
Each waypoint $w_m \in \mathcal W$ is then associated with a temporal region of the trajectory centered at its nominal time $\hat t_m$.
% which determines where along the trajectory the waypoint exerts influence, illustrated in Fig~\ref{fig:maze_trajectory}. 
Rather than enforcing hard interpolation through waypoints, we use them as a soft scaffold, allowing the diffusion process to produce locally plausible trajectories.

To this end, we define a triangular guidance window around each waypoint:
\[
\lambda_m(t)
=
\max\!\left(
0,\,
1 - \frac{|t-\hat t_m|}{r}
\right),
\]
where $r > 0$ is the window radius. This weight is maximal at $t = \hat t_m$ and decays linearly to zero with temporal distance. In order to conform with the existing overlap guidance term, for all our experiments we set the window size to match the overlap length $\mathcal{O}$.
The resulting waypoint guidance energy for a trajectory $\tau$ is
\[
E_{\mathcal W}(\tau)
=
\sum_{m=0}^{M}
\sum_{t=1}^{H}
\lambda_m(t)\,
\|s_t - w_m\|_2^2.
\]
% To inject the missing global structure, we leverage the waypoint sequence
% $\mathcal W$ via cost-based diffusion guidance~\citep{carvalho2023motion}.
% First, the nominal time of the final waypoint $w_M$ is treated as the nominal time of reaching the goal, which is used to initialize the noisy trajectory prior to the denoising process. Each waypoint \(w_m\in \mathcal W\) is then associated a poriton of the denoised trajectory  with the nominal trajectory time \(\hat t_m\), which determines which portion of the trajectory the waypoint should influence. 
% Rather than forcing the trajectory to rigidly interpolate waypoints, 
% we employ them as a soft scaffold and let diffusion determine locally-plausible realizations. To this end, we define a triangular guidance window around each waypoint
% \[\lambda_m(t)
% =
% \max\!\left(
% 0,\,
% 1 - \frac{|t-\hat t_m|}{r}
% \right),\]
% where \(r > 0\) is the window radius. This weight is maximal at the waypoint time \kiril{$\hat t_m$} and decays linearly to zero away from it, yielding, for a given trajectory $\tau$, the waypoint guidance energy \kiril{$M$ undefined}
% $E_{\mathcal W}(\tau)
% =
% \sum_{m=0}^{M}
% \sum_{t=1}^{H}
% \lambda_m(t)\,
% \|s_t - w_m\|_2^2$.
Which induces the guided distribution
\[
p_\theta(\tau \mid s_{\mathrm{start}}, s_{\mathrm{goal}}, \mathcal W)
\propto
p_{\theta}(\tau \mid s_{\mathrm{start}}, s_{\mathrm{goal}})
\exp\!\big(-E_{\mathcal W}(\tau)\big).
\]
At each denoising step, we augment the base denoising score by the gradient of the waypoint cost:
\[
\nabla_\tau \log p_\theta(\tau \mid q_s,q_g,\mathcal W)
\approx
\nabla_\tau \log p_\theta(\tau \mid q_s,q_g)
-
 \nabla_\tau \big(E_{\mathcal W}(\tau)\big).
\]
This form preserves the original compositional model while biasing denoising toward trajectories close to the graph solution.

\section{Experiments}
\label{experiments}
We evaluate the performance of XDiffuser to demonstrate that high-level extrinsic graph guidance improves long-horizon goal reaching. Moreover, we show that the same graph scaffold can be reused for unseen planning tasks (i.e., MAPF and inspection planning) by changing the high-level graph objective at test time. Ablation studies are discussed in Appendix~\ref{ablation}. %? Third, which design choices matter most in XDiffuser?  
%In all experiments, we use XDiffuser to generate a state trajectory, and during execution employ a separate inverse dynamics model or low-level controller. 
Our goal-reaching and MAPF experiments are conducted in the OGBench~\citep{ogbench_park2025} suite. For the bridge inspection experiment, we use a separate PyBullet simulation environment~\citep{pybullet}. 
% \adir{I think this paragraph needs clerification. Is it what is done, or a suggection for actual execution?}
All experiments were conducted on a single RTX 3090 GPU.

\subsection{Does extrinsic graph guidance improve long-horizon goal reaching?}
\label{ogbench}

We evaluate generation of long goal-reaching trajectories when the base diffusion model is trained on short demonstrations alone, illustrated in Fig.~\ref{fig:experimental_settings} . Moreover, we evaluate XDiffuser's performance based on the \textit{Explore} dataset, comprised of random-walk style demonstrations.

% We wish to measure XDiffuser's ability to sample successful trajectories after being trained on the suboptimal data of the  In all cases the diffusion model is trained only on short demonstrations, while successful test-time behavior may require trajectories an order of magnitude longer. \kiril{Here there's repetition w.r.t to previous sentence  It's also unclear why Stitch is emphasized for short demonstrations if it is stated here that all demostrations are short}

\textbf{Baselines}. We compare XDiffuser against the alternatives discussed in Section~\ref{related_work}. \textbf{CD}~\citep{luogenerative} represents pure compositional diffusion, and shares the same underlying diffusion model weights as XDiffuser, but relies entirely on local stitching without explicit search. \textbf{CDGS}~\citep{mishra2026cdgs} augments CD with a population-based search over candidate trajectories during denoising, reinforced by multiple resampling iterations. \textbf{C-MCTD}~\citep{yoon2025cmctd} places search directly inside denoising via tree search, spending inference-time compute on branching over partially denoised rollouts. Finally, \textbf{GAS}~\citep{baek2025graph} is a strong graph-based baseline that %is not diffusion-based, but instead 
uses a value-based policy to follow the shortest path over a pruned and clustered temporal distance graph. This makes GAS a useful comparison point for isolating the benefit of combining graph structure with a generative trajectory model. 

\textbf{Evaluation Setup}. We report \emph{execution success rate}: a rollout is successful if the robot reaches the goal position in the maze within the episode time limit. For each environment, we evaluate on five different start--goal tasks, each evaluated with 20 episodes, and the full evaluation is repeated over five random seeds. The mean and std are computed across seeds. All methods are evaluated on the same OGBench goal-reaching protocol, where the policy must execute the generated or planned trajectory in the environment rather than merely produce a geometrically-valid plan.  

\begin{table}[h]
\centering
\small
\begin{tabular}{lccccc}
\toprule
\textbf{Environment} & \textbf{CD} & \textbf{C-MCTD} & \textbf{CDGS} & \textbf{Ours} & \textbf{GAS} \\
\midrule
PointMaze (Stitch Giant)       & $68 \pm 3$  & $\mathbf{100 \pm 0}$ & $82 \pm 4$  & $\mathbf{100 \pm 0}$              & -- \\
\midrule
AntMaze (Stitch Giant)         & $65 \pm 3$ & $75 \pm 18$           & $84 \pm 3$  & $\mathbf{90.0 \pm 2.2}$   & $88.3 \pm 3.6$ \\
AntMaze (Stitch Large)         & $86 \pm 2$ & $94 \pm 9$            & --          & $90.6 \pm 2.2$            & $\mathbf{96.3 \pm 0.9}$ \\
AntMaze (Stitch Medium)        & $96 \pm 2$ & $98 \pm 6$            & --          & $93.0 \pm 3.0$            & $\mathbf{98.1 \pm 1.2}$ \\
\midrule
AntMaze (Explore Large)        & $27 \pm 1$ & --                    & --          & $\mathbf{98.5 \pm 0.5}$   & $94.2 \pm 3.0$ \\
AntMaze (Explore Medium)       & $81 \pm 2$ & --                    & --          & $\mathbf{99.3 \pm 0.5}$   & $94.2 \pm 3.0$ \\
\bottomrule
\end{tabular}
\vspace{3pt}
\caption{Success rate comparison (mean $\pm$ std) across OGBench environments. Results for other methods are taken from their respective papers. Results not reported are marked as blanks (--). }
\label{tab:results_no_ttgs}
\vspace{-1.5em}
\end{table}

\textbf{Results.} Table~\ref{tab:results_no_ttgs} shows XDiffuser substantially improves long-horizon goal reaching over pure CD. The improvement is most pronounced in settings where local stitching is insufficient: the \textit{Giant} maze and the \textit{Explore} datasets. On AntMaze \textit{Explore Large}, for example, CD succeeds only $27\%$ of the time, whereas XDiffuser reaches $98.5\%$ success. This gap suggests that when demonstrations are suboptimal and do not directly provide clean expert-like paths, relying only on probabilistic inference is insufficient. In contrast, a graph 
% provides a global abstraction of the state space, 
allows XDiffuser to identify promising waypoints before invoking the diffusion model. % for local trajectory generation.
Compared C-MCTD and CDGS, XDiffuser allocates inference-time search to a compact high-level graph rather than branching over partially denoised trajectories, which becomes increasingly important as the required horizon grows. 
% In long-horizon mazes, 
Branching directly within denoising can be expensive and unstable, since each partial rollout must remain dynamically plausible while also making progress toward a distant goal. 
% This may also explain the relatively high variance observed for C-MCTD, especially on AntMaze \textit{Stitch Giant}, where its performance varies substantially across seeds. An additional contrast between XDiffuser and intrinsic search diffusers is runtime. 
Moreover, denoising is \emph{time-consuming}---CDGS took on average $10\times$ longer (on our hardware) to generate, in accordance with the multiple resampling steps it performs. Since C-MCTD does not have a publicly available implementation, we could not measure its runtime on our hardware. However, on their much stronger hardware (8 NVIDIA RTX
4090 GPUs) they report runtimes which are $5\times$ longer than XDiffuser for Pointmaze \textit{Giant} on our setup. Additionally, as evident in their reporting, runtime scales exponentially as planning horizon grows due to their branching factor, indicating an even larger gap on more complex tasks.
% XDiffuser on the other hand leaves exploration to the graph algorithm, allowing diffusion model to focus on producing a single feasible trajectory. \kiril{consider removing this sentence. repetition of the first sentence}  

% \kiril{discuss here running times. important}

The results also reveal where graph guidance is less critical. On the shorter and easier \textit{Stitch Medium} and \textit{Stitch Large} tasks, the base CD model is already strong, reaching $96\%$ and $86\%$ success respectively. In these settings, waypoint guidance provides little improvement and can even slightly reduce performance when waypoints over-constrain an already reliable denoiser. This suggests that XDiffuser is most beneficial when the task requires genuine long-horizon planning. %, rather than when short-horizon composition alone is sufficient.

GAS provides an important comparison since it also uses a graph planner, but without a diffusion planner, 
% GAS performs very well on the shorter \textit{Stitch} tasks and 
and achieves the best results on AntMaze \textit{Stitch Medium} and \textit{Stitch Large}. However, XDiffuser is stronger on the most challenging \textit{Explore} settings and on AntMaze \textit{Stitch Giant}, highlighting the robustness of XDiffuser as problem horizon grows, as well as the advantage of fine-grained trajectory guidance which we evaluate in the following section. 

% These results suggest that graph structure alone is not always enough: combining high-level graph search with a reusable diffusion prior can improve robustness when data is suboptimal or the required route is especially long. Moreover, unlike purely graph-following methods, XDiffuser retains a generative trajectory model, enabling fine-grained trajectory synthesis and control beyond shortest-path goal reaching, which we evaluate in the following section.
% \YH{I will elaborate slightly more on advantages of diffusion vs value based in related work, and refer there} \kiril{good} 
% Since the original implementation is not publicly available, we could not properly reproduce these results and test our hypothesis. \adir{is it wise to say that?}
% GAS remains competitive on some AntMaze Stitch tasks, but our method is \YH{...TODO}.

% more modular: the same state-space diffusion prior can later be reused for different planning problems without retraining.
\vspace{-0.5em}
\subsection{Does extrinsic graph search generalize to unseen tasks?}
\label{unseen_tasks}
We test the modularity of the graph layer, by fixing the learned diffusion model and graph structure, but changing the high-level planning objective at test time. %If graphs provide a useful planning interface, then the same local generator should support qualitatively new behaviors beyond shortest-path goal reaching. 
We focus on MAPF and inspection planning as two representative unseen tasks; both require combinatorial reasoning not learned by the model, but which could be accommodated by graph algorithms.

% \subsubsection{Can the same framework solve multi-agent path finding?}
\subsubsection{Can search alone unlock zero-shot reasoning with diffusion?}
\label{MAPF}

We consider multi-agent path finding (MAPF) on AntMaze Stitch Medium with $n\in\{2,3,4\}$ ant robots, shown in Fig.~\ref{fig:experimental_settings}. Each agent $i$ is assigned a start-goal pair $(s^{i}_{\mathrm{start}}, s^{i}_{\mathrm{goal}})$, and the planner must output a collection of state trajectories $\{\tau^i\}_{i=1}^n$, where $\tau^i=(s^i_1,\dots,s^i_T)$. A joint plan is successful if every agent reaches its goal and all pairwise collision constraints are satisfied: for every pair $i \neq j$ and every timestep $t$, we require
$\|pos(s^i_t) - pos(s^j_t)\|_2 \ge \delta$,
with collision threshold $\delta$, where $pos$ is the position of the robot. This setting is intentionally out of distribution for a diffusion model trained only on single-agent data and never observes interaction constraints during training. We evaluate 20 episodes, over 3 random seeds each, where in each episode each agent executes one of the five queries from the single agent AntMaze environment. We report success rate where a success is defined by all agents reaching their destination with no collisions.

\textbf{Our method.} We instantiate XDiffuser with a multi-agent priority-planner~\citep{erdmann1987multiple} as \textbf{PP-XDiffuser}, in which each agent first solves a prioritized planning problem on its respective graph. Planning is still based on a shortest path, but each agent treats higher priority agents as dynamic obstacles on the graph. The resulting waypoint sequences are used to guide diffusion generation of each agent, as in the goal-reaching setting. 

% We evaluate three methods. \textbf{Naive CD} plans independently for each agent and ignores collisions with the others. \textbf{Guided CD} evaluates the effect of diffusion guidance without search. We instantiate multi-agent guidance as shown in (~\citep{shaoul2025multirobot}) which uses repulsion from other agents' trajectories to guide denoising of a pretrained single-agent diffusion planner. Unlike XDiffuser, this lacks any form of prior search. 
% \textbf{PP-XDiffuser} first solves a prioritized planning graph problem with reservations over graph states or edges to avoid future collisions, and then refines each agent's reserved waypoint sequence using the same single-agent diffusion planner as in our main method.

\textbf{Baselines}. We compare 4 other methods of utilizing the same pretrained single-agent model at test time. \textbf{Naive CD} and \textbf{Naive CDGS} plan independently for each agent using CD or CDGS, respectively, and ignore collisions. \textbf{Guided CD} evaluates the effect of diffusion guidance without search: following~\citet{shaoul2025multirobot}, it applies prioritized repulsion guidance during denoising, so that each agent is repelled only from the trajectories of higher-priority agents. This makes it the guidance-only analogue of prioritized MAPF, but without any explicit search over a separate graph structure. \textbf{Prioritized CDGS} augments CDGS's population-based search by penalizing trajectories in collision with higher priority agents. This setup isolates our main question: can explicit search 
% over a compact graph 
induce zero-shot multi-agent coordination from a purely single-agent diffusion prior? All methods share the same model weights, and denoise three segments, following~\citet{luogenerative}.

\begin{table}[h]
\centering
\small
\resizebox{\linewidth}{!}{
\begin{tabular}{llccccc}
\toprule
\textbf{Environment} & \textbf{\#Agents} & \textbf{Naive CD} & \textbf{Naive CDGS} & \textbf{Guided CD} & \textbf{Prioritized CDGS } & \textbf{PP-XDiffuser} \\
\midrule
\multirow{3}{*}{AntMaze Stitch Medium}
 & 2 & $38.33 \pm 6.24$ & $48.33 \pm 2.36$ & $40.00 \pm 4.08$ & $65.00 \pm 8.16$ & $\mathbf{ 88.33 \pm 6.24}$ \\
 & 3 & $3.33 \pm 4.71$  & $10.00 \pm 4.08$ & $15.00 \pm 10.80$ & $30.00 \pm 4.08$ & $\mathbf{73.33 \pm 2.36}$ \\
 & 4 & $0.00 \pm 0.00$  & $1.67 \pm 2.36$ & $1.67 \pm 2.36$ & $13.33 \pm 6.24$ & $\mathbf{58.33 \pm 13}$ \\
\bottomrule
\end{tabular}
}
\vspace{3pt}
\caption{Performance comparison %(mean $\pm$ standard deviation) 
on AntMaze Stitch Medium with varying numbers of agents.}
\label{tab:mapf_results}
\vspace{-1.0em}
\end{table}

\textbf{Results.} Table~\ref{tab:mapf_results} demonstrates that search is a crucial component for handling multi-agent coordination zero-shot. As the number of agents grows, naive single-agent plans fail frequently, and Guided CD's purely-local repulsive guidance is insufficient to prevent deadlocks and collisions. In contrast, only the methods that embed search, Prioritized CDGS and PP-XDiffuser, can coordinate $4$ agents. However, XDiffuser's extrinsic search offers a much more effective alternative % compared to CDGS's intrinsic search 
by first rapidly forming a coarse global plan which satisfies the new task constraints, and only then initiating the denoising process. As a result, for 4 agents XDiffuser achieves $58\%$ success rate compared with CDGS's $13\%$, as well as being more efficient---with prioritized search lasting $10$ seconds, and denoising $12\times4=48$ seconds, while CDGS generation takes $120\times 4=480$ seconds.

% \subsubsection{Can the same framework solve TPS-style reasoning tasks?}
\subsubsection{How well does extrinsic search scale in inspection planning?}
\label{inspection_plan}

% \begin{wrapfigure}{r}{0.45\linewidth}
% \vspace{-1.0em}
%     \centering
%     \includegraphics[width=1\linewidth]{assets/POI_vs_time_uncertainty.pdf}
%     \vspace{-1.0em}
%     \caption{Enter Caption}
%     \label{fig:placeholder}
%     \vspace{-1.0em}
% \end{wrapfigure}

% We consider an inspection planning (IP) problem~\citep{fu2019toward}, in which a robot equipped with a sensor must compute a minimum-cost trajectory $\tau$ that enables the observation of a set of points of interest (POIs) $\mathcal{P}=\{p_1, \dots, p_k\}$ throughout the environment.
We consider an inspection planning (IP) problem~\citep{fu2019toward}, in which a drone equipped with a sensor is tasked with observing a set of points of interest (POIs) $\mathcal{P}=\{p_1, \dots, p_k\}$ throughout the environment.
While superficially related to multi-goal reaching, IP differs fundamentally in that POIs need not be physically reached %---indeed, they are often embedded within obstacles---
but rather observed from feasible vantage points. Consequently, the problem requires jointly optimizing (i) the inspection order of POIs, (ii) the selection of observation viewpoints for each POI, and (iii) the connecting motion between viewpoints. %, in order to synthesize an efficient trajectory.

% \YH{Experimental setting}
% In our experiments, we instantiate an IP task where a simulated drone robot inspects a large scale bridge as shown in Fig~\ref{bridge}, with $n\in{4,8,16,64,128}$ POIs. The drone is defined with $3D$ linear dynamics, with its state including position and velocity, and its actions defining the acceleration in the $\{x,y,z\}$ axes. We provide further details on data collection and training in Appendix~\ref{ip_experiment}.
% In order to focus on the combinatorial aspect of IP, we apply to all methods two relaxing assumption: first, we assume perfect test-time tracking, and treat the planned trajectory as the de-facto executed trajectory. Second, we allow minor collisions - i.e., up to $1$ second in collision, but otherwise terminate an episode when the drone is in collision for a longer period. 

In our experiments, we consider a inspection of a large-scale bridge by a flying drone (Fig.~\ref{fig:experimental_settings}) with $n \in \{4, 8, 16, 64, 128\}$ POIs distributed over the bridge structure using farthest-point sampling over mesh vertices. The drone obeys $3$D linear dynamics, where the state consists of position and velocity, and the actions correspond to accelerations along the $x$, $y$, and $z$ axes. As a shared base model, we train a diffusion model with the same hyperparameters as used for PointMaze Giant.
Full details on data collection, model training and problem formulaion are provided in Appendix~\ref{app:inspection_planning}.

To focus on the combinatorial aspects of IP, we apply two simplifying assumptions across all methods. First, we assume perfect tracking at test time and treat the planned trajectory as the executed trajectory. Second, we allow minor collisions, defined as up to one second of contact; episodes are terminated if a collision persists longer than this threshold.
For each $n$ POIs we simulate three episodes each starting at different start location, and repeat each episode over three random seeds. We evaluate each method according to the achieved \textit{coverage}---percentage of POIs observed during execution, where an observation is registered once the drone is within some threshold Euclidean distance from the POI.

\begin{figure}
    \centering
\includegraphics[width=0.9\linewidth]{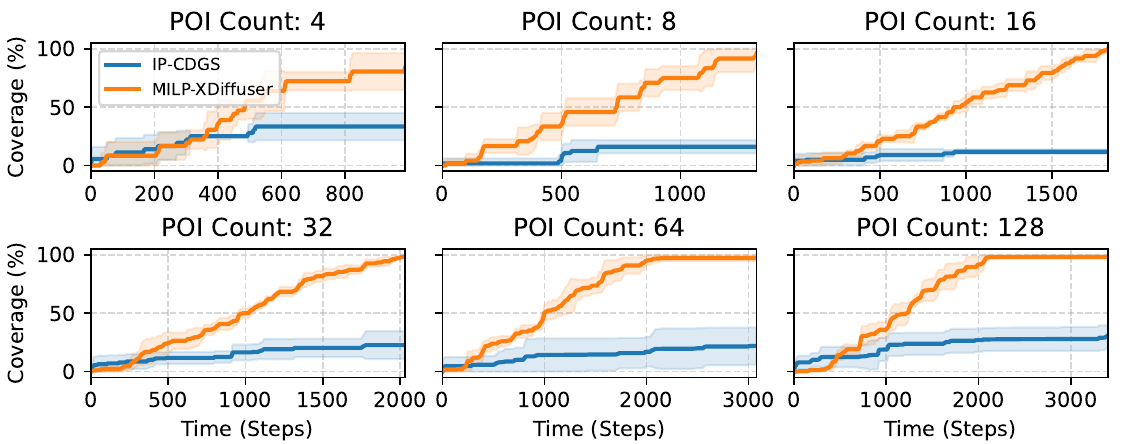}
    \caption{POI coverage over mission time for the inspection-planning task.}
    \label{fig:PI_coverage}
\end{figure}

\textbf{Our method.} We adapt XDiffuser to inspection planning (IP) by decomposing the problem into high-level combinatorial planning and low-level trajectory generation. For each POI, we associate a set of candidate viewpoints by selecting its $K$ nearest neighbors in the state space and augmenting the graph accordingly (Section~\ref{sec:graph_construction}). We then apply the mixed-integer linear programming (MILP) Graph-IP solver of~\citet{morgan2026scalable} on the resulting graph to obtain a sequence of vertices forming a valid covering tour over POI viewpoints. This sequence defines a sparse set of waypoints, which we use to guide the diffusion model toward generating a dynamically feasible inspection trajectory. We refer to this instantiation as MILP-XDiffuser.

% We adapt XDiffuser's graph routine to IP by leveraging the mixed-integer linear-program (MILP) based solver by~\citet{morgan2026scalable}, which returns a sequence of vertices constituting a valid tour between POIs. We term this variant MILP-XDiffuser. At test time, for each POI we select the $K$ nearest neighbors and append each to XDiffuser's graph as described in~\ref{sec:graph_construction}

\textbf{Baseline.} As an intrinsic-search baseline, we adapt CDGS by modifying its search objective to favor inspection coverage by ranking candidate trajectories according to the number of POIs observed, encouraging generation of high-coverage solutions. For a fair comparison, we set the number of CDGS's trajectory segments to match the number implied by the MILP-XDiffuser plan.

%
%we employ as a search mechanism the graph-based IP solver , which uses a flow-based branch-and-cut formulation with anytime guarantees. % and strong scalability. 
% While constructing the temporal distance graph 
% $\mathcal G=(\mathcal V,\mathcal E)$, and associate each vertex $v \in V$ with the subset of POIs visible from it via a 
% visibility function, which we choose as a threshold over Euclidean distance. 
% $\chi: V \rightarrow 2^{\mathcal{P}}$
% IP is then solved over this graph, producing a sequence of inspection states $\tau=(x_1,\dots,x_T)$ that ensures full POI coverage at minimum cost. These inspection states are subsequently connected using our pretrained compositional diffusion model. %, which generates dynamically feasible trajectories between successive states. 
% This decomposition highlights the key advantage of our approach: the graph layer resolves the combinatorial inspection-planning problem, while the diffusion model synthesizes locally feasible motion between inspection states.

% \textbf{Results.}
% Fig.~\ref{fig:PI_coverage} illustrates how coverage evolves over execution time for each method, with final coverage summarized in Table~\ref{tab:ip_coverage}. MILP-XDiffuser consistently achieves near-complete coverage, exceeding $98\%$ on instances with 16 POIs or more, while exhibiting a steady and monotonic increase in coverage throughout execution. In contrast, IP-CDGS attains modest early gains but fails to sustain progress, plateauing well below full coverage and never exceeding $50\%$.

\begin{table}[t]
\centering
\small
\begin{tabular}{lccccccc}
\toprule
Method & 4 & 8 & 16 & 32 & 64 & 128 & Avg. \\
\midrule
IP-CDGS & 33.3$\pm$12.1 & 12.5$\pm$8.6 & 11.8$\pm$2.0 & 20.1$\pm$13.7 & 22.0$\pm$16.4 & 31.0$\pm$10.2 & 21.8 \\
MILP-XDiffuser & {83.3$\pm$12.1} & {95.8$\pm$6.1} & {100.0$\pm$0.0} & {98.6$\pm$1.6} & {97.2$\pm$2.9} & {98.3$\pm$1.8} & {95.5} \\
\bottomrule
\end{tabular}
\vspace{3pt}
\caption{Final POI coverage (\%) for the inspection-planning task across different numbers of POIs.}
\label{tab:ip_coverage}
\vspace{-2.5em}
\end{table}

% While both methods share the underlying pretrained model, MILP-XDiffuser consistently achieves higher coverage by forming well articulated tours over the inspection graph. In contrast, CDGS's intrinsic search tends to exhibit myopic behavior, achieving early gains in coverage but failing to complete the inspection task. This gap becomes more pronounced in larger instances, demonstrating that structured planning plays a key role for scaling diffusion-based approaches to complex planning problems.
% These results show the effectiveness of pairing diffusion models with extrinsic planners for long-horizon planning, especially as combinatorial search space grows larger.

\textbf{Results.} 
Fig.~\ref{fig:PI_coverage} illustrates how coverage evolves over execution time for each method, with final coverage summarized in Table~\ref{tab:ip_coverage}. Although both methods use the same pretrained diffusion model, MILP-XDiffuser consistently achieves near-complete coverage, exceeding $95\%$ on instances with 8 POIs or more, while exhibiting a steady and monotonic increase in coverage throughout execution. 
In contrast, IP-CDGS exhibits myopic search behavior, with modest early gains but fails to sustain progress, plateauing well below full coverage and never exceeding $50\%$. These results highlight the benefit of dedicated extrinsic planning for scaling diffusion planners to long-horizon and complex combinatorial problems. 

\vspace{-0.5em}
\section{Discussion and Conclusion}
\textbf{Limitations.} %While our approach improves long-horizon planning by combining graph search with diffusion, several limitations remain. First, our method abides the fundamental constraints of offline learning. Both the graph and the diffusion model rely on states covered in the dataset. As a result, the planner cannot reliably generate behavior that lies far outside the dataset's support. Although the graph can propose long-range compositions, these are only meaningful when intermediate connectivity is sufficiently represented; in sparse or biased datasets, the planner may fail to discover feasible global routes. Second, execution quality depends on the downstream control mechanism. Similar to past work~\citep{luogenerative}, in our experiments failures often occur due to the inverse dynamics model, particularly in environments with tight constraints or contact-rich dynamics. While diffusion improves the smoothness and plausibility of trajectories compared to raw graph paths, it does not explicitly account for execution uncertainty, which can lead to collisions or tracking errors at deployment.
Our graph construction is intentionally simple, relying on sampled states and temporal distance connectivity. While this suffices for the tasks we consider, it may prove to be a hurdle in more complex settings, e.g. with stochastic dynamics. It may also hinder performance when the sampled graph is disconnected, as shown in~\ref{graph_ablation}. Moreover, as temporal distance is symmetric, it requires the graph to be undirected, which does not faithfully capture many robotic systems. 
Incorporating richer graph representations, learned abstractions, or uncertainty-aware connectivity remains an open direction. We look forward to exploring formulations that better align with the diffusion model's inference, e.g., learn likelihood as graph edge costs, as proposed by~\citet{liu2020hallucinative}.
% , mirroring the diffusion model's liklihood maximization. 

% Additionaly, while our diffusion gradient-guidanceis simple and training free, it could ignore important context, e.g., waypoints located just behind an obstacle might harm the resulting trajectory. Incorporation of training-based guidance might help improve performance in some complex environments.
% capture the capabilities of the test-time diffusion model, rather than relying on a surrogate as temporal distance. 

%Finally, although our method is more efficient than search-in-denoising approaches, even a single denoising process still incurs nontrivial computational cost compared to value-based policies. \kiril{really? Dijkstra takes longer than denoising?} \YH{No, Dijkstra+Diffusion takes longer than smaller RL policies} This may limit applicability in real-time or resource-constrained settings without further optimization. \YH{(but maybe I'll have time to test the graph-warmup idea)}

\textbf{Conclusion.}
We studied the problem of long-horizon offline planning from short, suboptimal trajectory data. We identified a key limitation of compositional diffusion and planning-as-inference approaches: when long-horizon behavior is never observed during training, local generative consistency is insufficient to ensure globally coherent plans. To address this, we proposed a simple decomposition: use graph search for global coordination and diffusion for local trajectory generation. Our method XDiffuser constructs a graph over offline states, plans a sparse sequence of waypoints via search, and then guides a pretrained compositional diffusion model to synthesize a continuous, dynamically plausible trajectory. By placing search outside the denoising loop, the approach is both efficient and modular, enabling adaptive horizons and flexible test-time objectives.
Empirically, we showed that this combination leads to strong improvements on the OGBench suite, consistently outperforming prior diffusion-based planners and matching or exceeding state-of-the-art value-based methods, particularly in the low-quality-data regime. Moreover, the same framework extends naturally to more complex tasks such as multi-agent path finding and combinatorial routing, simply by changing the high-level graph objective.

{
\small
\bibliographystyle{unsrtnat}
\bibliography{refs}
}

% [1] Alexander, J.A.\ \& Mozer, M.C.\ (1995) Template-based algorithms for
% connectionist rule extraction. In G.\ Tesauro, D.S.\ Touretzky and T.K.\ Leen
% (eds.), {\it Advances in Neural Information Processing Systems 7},
% pp.\ 609--616. Cambridge, MA: MIT Press.

% [2] Bower, J.M.\ \& Beeman, D.\ (1995) {\it The Book of GENESIS: Exploring
%   Realistic Neural Models with the GEneral NEural SImulation System.}  New York:
% TELOS/Springer--Verlag.

% [3] Hasselmo, M.E., Schnell, E.\ \& Barkai, E.\ (1995) Dynamics of learning and
% recall at excitatory recurrent synapses and cholinergic modulation in rat
% hippocampal region CA3. {\it Journal of Neuroscience} {\bf 15}(7):5249-5262.
% }

%%%%%%%%%%%%%%%%%%%%%%%%%%%%%%%%%%%%%%%%%%%%%%%%%%%%%%%%%%%%

\appendix

\section{Appendix}

\subsection{Ablation study and design choices}
\label{ablation}
%\adir{a confusing name for ablation studies?}
Unless otherwise noted, ablations are performed on the AntMaze-Large-Stitch task.

\begin{wraptable}{r}{0.48\linewidth}
\vspace{-10pt}
\centering
\small
\begin{tabular}{lcccc}
\toprule
\textbf{n} & \textbf{k=10} & \textbf{k=20} & \textbf{k=30} & \textbf{k=40} \\
\midrule
500  & $\mathbf{0 \pm 0}$ & $90 \pm 9$ & ${95 \pm 0}$ & $90 \pm 0$ \\
750  & $\mathbf{0 \pm 0}$ & $90 \pm 13$ & ${93 \pm 8}$ & $88 \pm 3$ \\
1000 & $90 \pm 5$ & $83 \pm 3$ & ${95 \pm 0}$ & $90 \pm 13$ \\
1500 & $87 \pm 10$ & $82 \pm 14$ & ${92 \pm 3}$ & $90 \pm 9$ \\
\bottomrule
\end{tabular}
\vspace{-5pt}
\caption{Success rate across number of graph nodes $n$ and connectivity $k$.}
\label{tab:exp1}
\vspace{-10pt}
\end{wraptable}

\subsubsection{How sensitive is XDiffuser to graph size and connectivity?}
\label{graph_ablation}

% We jointly vary the number of sampled graph vertices and the graph connectivity parameter $k$, i.e., the number of nearest neighbors used when constructing the graph. This ablation tests whether our method depends on a finely tuned graph or merely on recovering coarse global connectivity. We find the method is relatively robust to both quantities: once the graph is dense enough to capture the main corridors of the state space, performance changes only mildly across a wide range of vertex counts and neighborhood sizes. Failures appear mainly when the graph becomes too small or too sparse to represent feasible long-range routes.

We jointly vary the number of sampled vertices \(n\) and the graph connectivity parameter \(k\) (number of nearest neighbors) to assess sensitivity to graph construction. Table~\ref{tab:exp1} shows a clear threshold behavior: for low connectivity (\(k=10\)), performance is poor unless the graph is very dense, indicating failure to capture long-range feasible paths. However, once connectivity increases (\(k \geq 20\)), success rates rise sharply and quickly saturate, remaining high across a wide range of \(n\). In particular, performance is consistently strong for \(k=30\), with low variance, suggesting reliable recovery of global routes. Beyond moderate coverage, increasing \(n\) yields diminishing returns, indicating that the method depends primarily on achieving sufficient connectivity rather than precise graph tuning.

\subsubsection{How sensitive is XDiffuser to the downsampling waypoint interval \texorpdfstring{$\bm{\Delta t}$}{dt}?}
% \subsubsection{How sensitive is XDiffuser to trajectory horizon time and the waypoint interval \texorpdfstring{$\bm{\Delta t}$}{dt}?}
\label{diffusion_ablation}

\begin{table}[h]
\centering
\small
\begin{tabular}{lcccc}
\toprule
{\texorpdfstring{$\bm{\Delta t}$}{dt}} & 8.0 & 12.0 & 24.0 & 48.0 \\
\midrule
{Success rate} & ${93 \pm 3}$ & \textbf{$95 \pm 0$} & $88 \pm 8$ & $92 \pm 6$ \\
\bottomrule
\end{tabular}
\caption{Success rate across different waypoint intervals $\bm{\Delta t}$.}
\label{tab:exp2_wp1}
\end{table}
We vary the waypoint downsampling interval $\Delta t$, maintaining a waypoint every $\Delta t$ steps along the planned trajectory. To evaluate the impact of this choice, we run experiments on the first 20 episodes of AntMaze Large, repeating each episode across five random seeds. The results indicate a modest trade-off: smaller $\Delta t$ yields denser supervision but may overconstrain the denoising process, while larger $\Delta t$ leads to overly sparse waypoints which provide weak guidance. Empirically, $\Delta t = 12$ achieves the best performance, and we adopt this value across all experiments.

\begin{figure}
    \centering
    \includegraphics[width=0.5\linewidth]{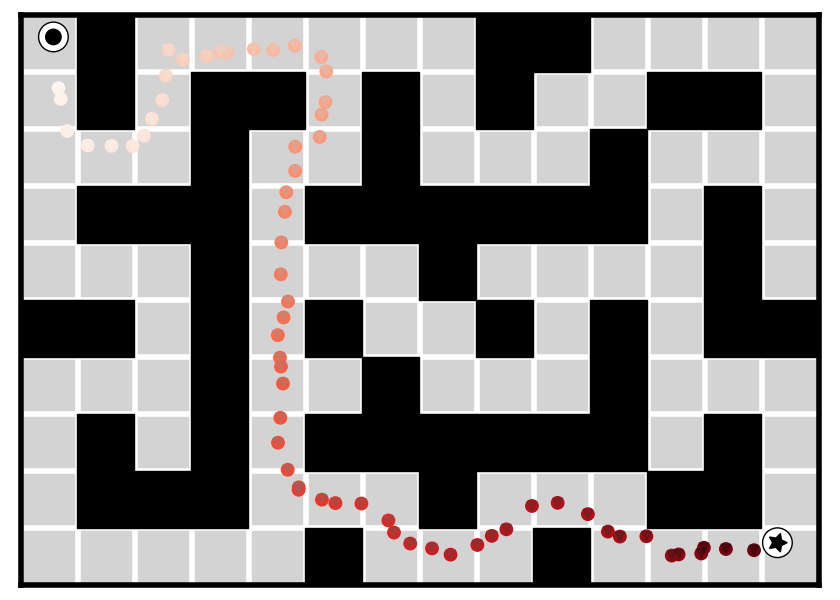}
    \caption{An example XDiffuser graph shortest path, prior to downsampling. Initial state is marked with a black circle, and goal with a star.}
    \label{fig:graph_path}
\end{figure}

\begin{figure}[t]
    \centering
    
    \begin{subfigure}[t]{0.4\textwidth}
        \centering
        \includegraphics[width=\linewidth]{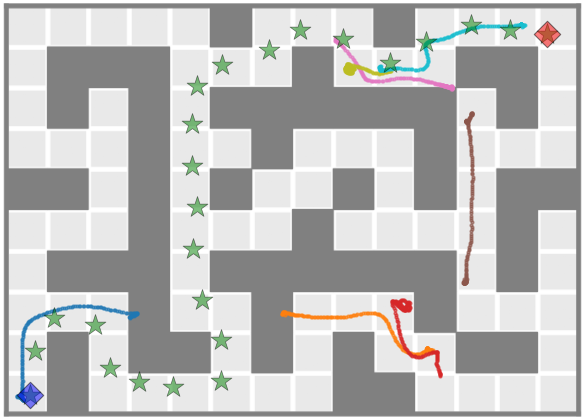}
        % \caption{Goal reaching.}
        \label{fig:bad_composition}
    \end{subfigure}
    % \hfill
    \begin{subfigure}[t]{0.4\textwidth}
        \centering
        \includegraphics[width=\linewidth]{assets/good_composition.png}
        % \caption{Multi-agent planning.}
        \label{fig:good_composition}
    \end{subfigure}
    
    \caption{\textbf{Guidance window effect.} During graph-guided generation every waypoint attracts states from the generated segments around its nominal time. Left: attracting a single states produces very weak guidance, and as a results segments adhere to their local denoising objective while ignoring the global waypoint structure. Right: by using a triangular guidance window, guidance is distributed along the trajectory creating effective guidance which properly aligns all segments.
     \ref{experiments}. 
    % \kiril{If space becomes an issue, embed the subcaption in a white text box within each figure and remove the subcaption}
    % Planning tasks evaluated in our experiments. Left: The agent represented by the orange ball plans a trajectory to the goal marked with a red star, using diffusion guided by waypoins marked as green stars. Middle: a multi agent planning scenario, where agents must reach their goal while avoiding collision with the others. Right: trajectory generated by XDiffuser for a bridge inspection  task featured in experiment~\ref{ip_experiment}. 
    }
    \label{fig:guidance_window}
    \vspace{-0.5em}
\end{figure}

% XDiffuser initilizes a noisy trajectory based on the nominal time path duration deduced from the waypoints sequence found during search.
% We multiply this duration by a time-dilation factor $\gamma$ to evaluate XDiffuser's sensitivity to temporal distance prediction errors, and test over 20 episodes, each repeated for 5 seeds.
% % that converts graph travel time into the diffusion horizon, $H_i = \lceil \gamma \hat{h}_i \rceil$. 
% As observed in compositional diffusion planning~\citep{luogenerative}, horizons that are too short force neighboring subtrajectories to satisfy overlap constraints aggressively, which can make them cut across invalid regions. In our setting, this manifests as trajectories that clip corners or pass through walls. Conversely, horizons that are too long create unnecessary detours and can induce a zig-zag walking pattern that increases the chance of colliding with walls and corners. This ablation clarifies that adaptive horizons are important, but they still require an appropriate amount of temporal slack.

\subsubsection{Is diffusion necessary after graph search?}
We compare the full method against a graph-only execution baseline that feeds the shortest path directly to the inverse dynamics model, without diffusion refinement, demonstrated in Fig~\ref{fig:graph_path}. Given a sequence of graph waypoints, we linearly interpolate between them to produce a continuous trajectory for execution. Somewhat surprisingly, this simple approach already achieves a \(77 \pm 1.4\%\) success rate, indicating that the learned graph provides a strong backbone for high-level planning.

However, its limitations become pronounced in longer-horizon settings. While interpolated waypoints can approximate a smooth trajectory, they often skim obstacles or introduce sharp turns that are difficult to execute robustly. These small tracking errors accumulate over time and are hard to recover from, leading to a sharp drop in performance. In particular, on AntMaze Giant, the graph-only baseline achieves just \(6 \pm 0\%\) success. This highlights the necessity of diffusion refinement for producing smooth, dynamically feasible trajectories that remain robust over long horizons.

\section{Inspection Planning using XDiffuser}
\label{app:inspection_planning}
This appendix provides additional details for the inspection-planning experiment described in Sec.~\ref{inspection_plan}. We describe the $3$D bridge environment, the drone dynamics, the construction of the offline trajectory dataset, the inspection graph used by MILP-XDiffuser, and the evaluation protocol.

\begin{figure}[h]
    \centering
    \begin{subfigure}[t]{0.48\linewidth}
        \centering
        \includegraphics[width=\linewidth]{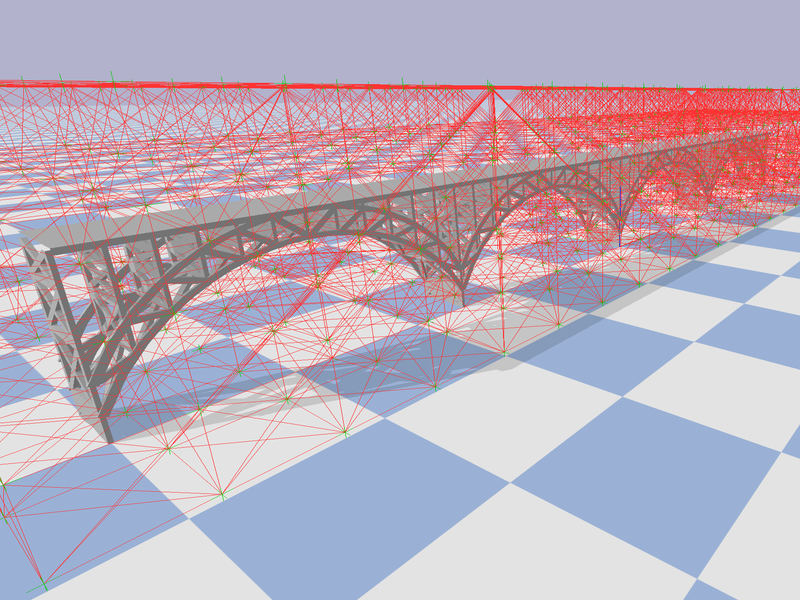}
        \caption{Workspace discretization as a $3$D planning graph with collision-free edges.}
        \label{fig:grid}
    \end{subfigure}
    \hfill
    \begin{subfigure}[t]{0.48\linewidth}
        \centering
        \includegraphics[width=\linewidth]{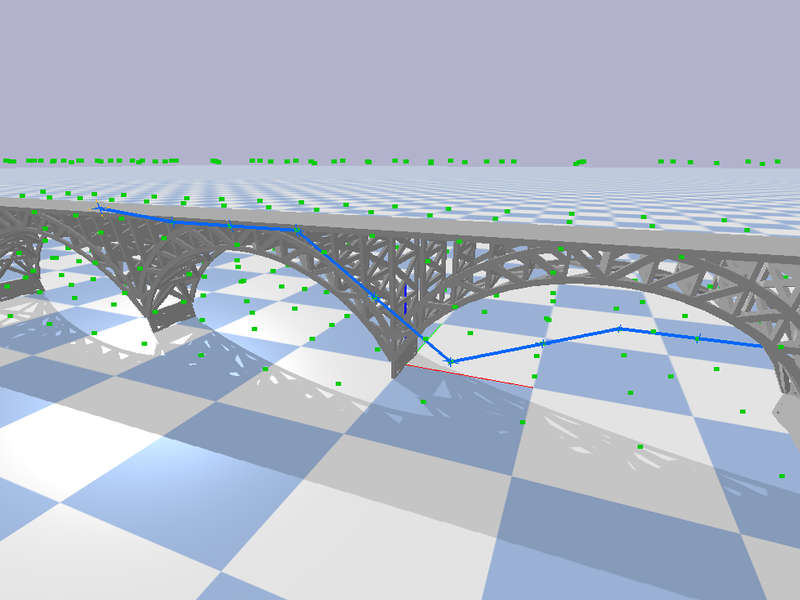}
        \caption{Geometric path generated on the sampled grid between two randomly sampled positions.}
        \label{fig:geom_path}
    \end{subfigure}

    \vspace{0.5em}

    \begin{subfigure}[t]{0.48\linewidth}
        \centering
        \includegraphics[width=\linewidth]{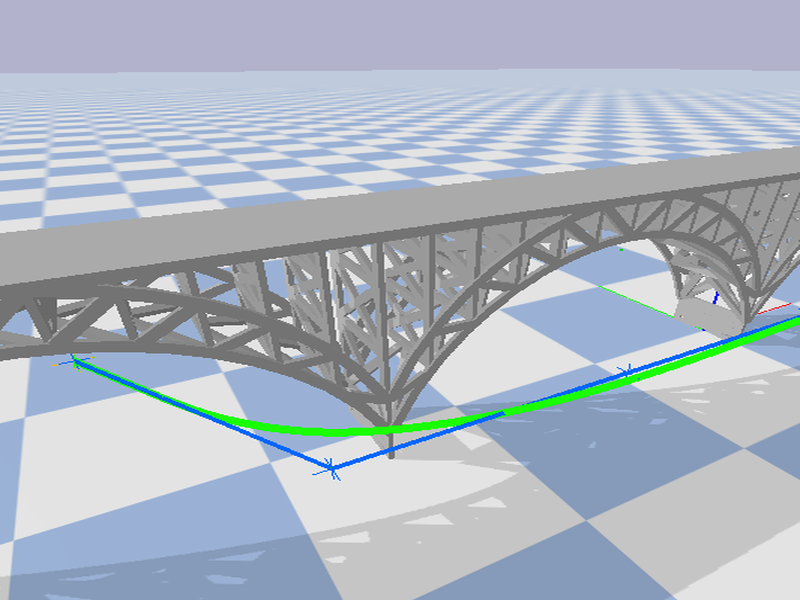}
        \caption{Dynamically feasible trajectory obtained via a realistic high-level drone controller, tracking rolling trajectory waypoints.}
        \label{fig:dyn_traj}
    \end{subfigure}
    \hfill
    \begin{subfigure}[t]{0.48\linewidth}
        \centering
        \includegraphics[width=\linewidth]{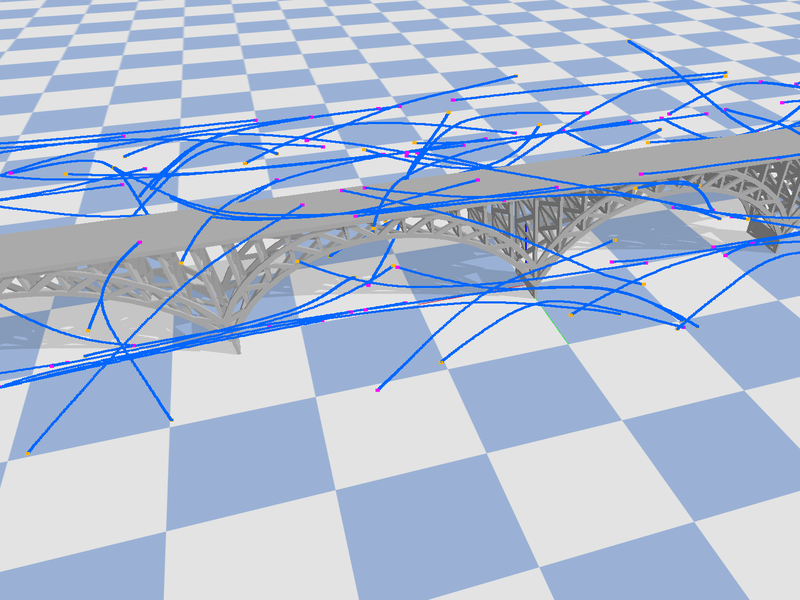}
        \caption{Collection of sampled trajectories forming the training dataset.}
        \label{fig:dataset}
    \end{subfigure}

    \caption{Dataset generation pipeline. 
    % (Top-left) The environment is discretized into a collision-free grid. 
    % (Top-right) Short geometric paths are generated over the graph. 
    % (Bottom-left) Paths are tracked under robot dynamics to obtain feasible trajectories. 
    % (Bottom-right) Repeated sampling yields a dataset of state-action trajectories.
    }
    \label{fig:dataset_pipeline}
\end{figure}

\subsection{Offline Dataset Construction}
\label{app:inspection_dataset}
We construct an offline dataset of short dynamically feasible drone trajectories in the bridge environment. 
The dataset is used only to train the local diffusion trajectory model; it does not contain demonstrations of the inspection-planning task, inspection rewards, or complete tours over POIs. This mirrors the long-horizon setting considered in the main experiments, where the model is trained on local behavior but evaluated on generalization tasks requiring global coordination.

\paragraph{Environment.}
We use a $3$D bridge model represented by a triangular mesh and simulate robot--environment interactions in PyBullet~\citep{pybullet}. The workspace is obtained from the bridge mesh bounding box, expanded by a fixed margin to allow free-space motion around the structure. All physical sizes are expressed in the normalized coordinate frame of the mesh.

\paragraph{Geometric path generation.}
To generate candidate motions, we first discretize the collision-free workspace into a $3$D grid with resolution $\rho=0.25$. Collision checking is performed directly against the mesh using a spherical drone body of radius $r=0.15$.
Grid vertices correspond to collision-free positions, and edges connect neighboring vertices under a $18$-connected neighborhood when the straight-line segment between them is collision-free, with $5$ collision-check samples per edge. 
For each dataset trajectory, we generate random collision-free start and goal positions, connect them to the motion planning grid via their six closest nearest grid vertices, and compute a shortest path on this grid using the A$^\ast$ algorithm with Euclidean distance heuristic.

\paragraph{Dynamics and tracking.}
Each geometric path is converted into a dynamically feasible trajectory using a PID controller applied independently along each axis under a double-integrator drone model. The state is $s_t=(p_t,v_t)\in\mathbb{R}^6$, where $p_t\in\mathbb{R}^3$ is position and $v_t\in\mathbb{R}^3$ is velocity, and the action $a_t\in\mathbb{R}^3$ specifies acceleration. We use the discrete-time dynamics
\[
p_{t+1}=p_t+\Delta t\,v_t,\qquad v_{t+1}=v_t+\Delta t\,a_t,
\]
with timestep $\Delta t=0.05\,\mathrm{s}$ and acceleration clipped to $\|a_t\|_\infty \leq 3$. This dynamics model captures the high-level behavior of a drone while abstracting away platform-specific actuation details, which are assumed to be handled by an inner control loop. The controller tracks the geometric path by selecting accelerations toward a rolling target waypoint. After rollout, each trajectory is clipped to $H_{\mathrm{train}}=200$ state-action pairs to match the OGBench \textit{Stitch} format~\citep{ogbench_park2025}, and is collision-checked again under the executed dynamics; trajectories that collide with the bridge are discarded.

\paragraph{Dataset construction and model training.}
Repeating the sampling, geometric planning, tracking, and validation procedure yields a dataset of $5{,}000$ state-action trajectories. The dataset contains only short point-to-point motions and does not include POIs, inspection rewards, or complete inspection tours. To avoid directional bias in the generated motions, we sample trajectories within a cubic bounding box rather than using the elongated bounding box of the bridge mesh. We use the state sequences to train the compositional diffusion model, using the same architecture, denoising objective, and optimization settings as in the main experiments. The same dataset is also used to construct XDiffuser's extrinsic guidance graph. At test time, inspection-specific structure enters only through the graph-level inspection planner, not through the diffusion training data.

\subsection{Inspection-Planning Experiment}
\label{app:inspection_experiment}
We now describe how the trained model is used to solve inspection-planning tasks. Each inspection instance is defined by a set of $n\in\{4,8,16,32,64,128\}$ points of interest (POIs) sampled on the bridge surface using farthest-point sampling. A POI is considered observed once the drone comes within distance $r_{\mathrm{obs}}=1$ of it.

To connect the POIs to XDiffuser's extrinsic guidance graph, we define an inspection relation between POIs and graph vertices based on spatial proximity. Specifically, for each POI, we associate a set of candidate viewpoint vertices from the graph. This inspection relation defines which graph vertices can observe which POIs.

Given the POI set and the inspection relation, we construct a graph-based inspection-planning problem over the same extrinsic graph used by XDiffuser. We solve this discrete problem using the Graph-IP planner of~\citet{morgan2026scalable}, which returns a sequence of graph vertices forming a covering tour: the selected vertices collectively observe all POIs while minimizing travel cost over the graph. Specifically, we use the single-commodity-flow MILP formulation, which is well suited to problem instances at this scale, with a timeout of three minutes. 

This high-level tour is then converted into temporal waypoints using cumulative graph edge costs, following the waypoint construction described in Sec.~\ref{method}. Finally, these waypoints are provided as guidance to the compositional diffusion model, which generates a continuous, dynamically feasible trajectory in the drone state space. We refer to this inspection-planning instantiation as MILP-XDiffuser.

\begin{figure}[h]
    \centering
    \begin{subfigure}[t]{0.48\linewidth}
        \centering
        \includegraphics[width=\linewidth]{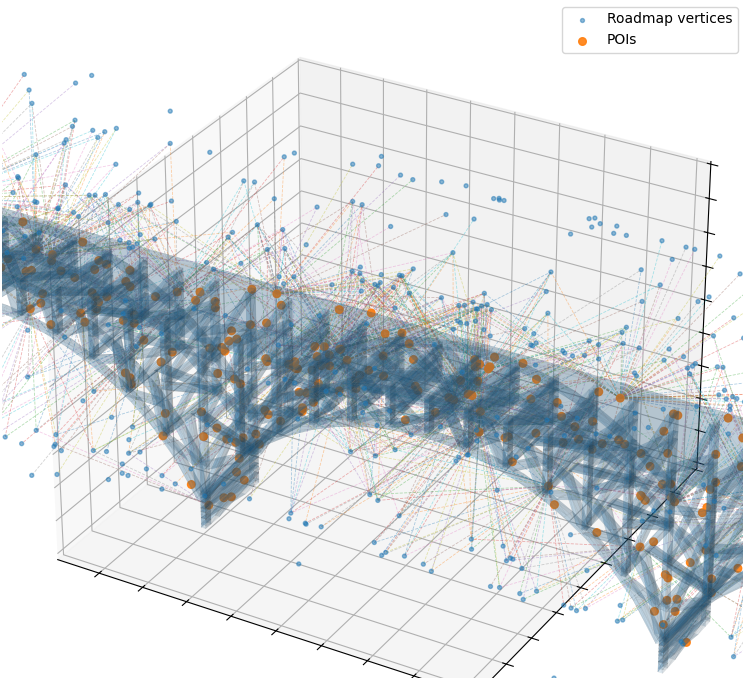}
        % \caption{Inspection instance with sampled POIs and their relation to roadmap vertices.}
        \label{fig:inspection_instance}
    \end{subfigure}
    \hfill
    \begin{subfigure}[t]{0.48\linewidth}
        \centering
        \includegraphics[width=\linewidth]{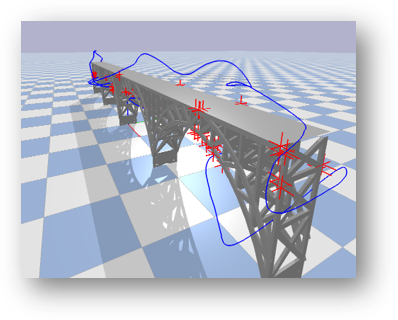}
        % \caption{Generated inspection trajectory guided by the high-level inspection plan.}
        \label{fig:inspection_trajectory}
    \end{subfigure}

    \caption{Inspection planning with XDiffuser. 
    (Left) POIs are sampled on the bridge surface and associated with nearby roadmap vertices through the inspection relation. 
    (Right) The graph-level inspection plan is provided as guidance to XDiffuser, which generates a dynamically feasible inspection trajectory.}
\label{fig:inspection_planning_pipeline}
\end{figure}

% \adir{Note to self - refer to our particular experiment, and add specific numbers}
% \adir{Writing is rigid and fregmented, improve}

% \adir{revisit figures}

% \newpage
% \input{checklist.tex}

\end{document}